\documentclass[letterpaper, 10 pt, conference]{ieeeconf}  
\IEEEoverridecommandlockouts                              

\usepackage[T1]{fontenc}  
\usepackage{cite}

\usepackage{amsmath,amssymb,amsfonts}
\usepackage{algorithmic}
\usepackage{graphicx}
\usepackage{textcomp}
\usepackage{xcolor}
\usepackage{courier}

\usepackage{verbatim}
\usepackage{booktabs}
\usepackage{multirow}
\usepackage{colortbl}
\usepackage{graphicx}
\usepackage{subfigure}
\usepackage{makecell}
\usepackage{txfonts}

\title{\LARGE \bf
Cross-Modality Time-Variant Relation Learning\\for Generating Dynamic Scene Graphs}

\author{Jingyi Wang, Jinfa Huang, Can Zhang, and Zhidong Deng$^{*}$
\thanks{Jingyi Wang is with Department of Computer Science, Tsinghua University, Beijing 100084, China (email: wang-jy20@mails.tsinghua.edu.cn)}%
\thanks{Jinfa Huang and Can Zhang are with the School of Electronic and Computer Engineering, Peking University, China (emails: jinfahuang@stu.pku.edu.cn, zhangcan@pku.edu.cn)}
\thanks{$^{*}$Zhidong Deng is with Beijing National Research Center for Information Science and Technology (BNRist), THUAI, Department of Computer Science, State Key Laboratory of Intelligent Technology and Systems, Tsinghua University, Beijing 100084, China (email: michael@tsinghua.edu.cn)}
}

\begin{document}

\maketitle
\thispagestyle{empty}
\pagestyle{empty}

\begin{abstract}
Dynamic scene graphs generated from video clips could help enhance the semantic visual understanding in a wide range of challenging tasks such as environmental perception, autonomous navigation, and task planning of self-driving vehicles and mobile robots. In the process of temporal and spatial modeling during dynamic scene graph generation, it is particularly intractable to learn time-variant relations in dynamic scene graphs among frames. 
In this paper, we propose a Time-variant Relation-aware TRansformer (TR$^2$), which aims to model the temporal change of relations in dynamic scene graphs. 
Explicitly, we leverage the difference of text embeddings of prompted sentences about relation labels as the supervision signal for relations. In this way, cross-modality feature guidance is realized for the learning of time-variant relations. 
Implicitly, we design a relation feature fusion module with a transformer and an additional message token that describes the difference between adjacent frames. 
Extensive experiments on the Action Genome dataset prove that our TR$^2$ can effectively model the time-variant relations. TR$^2$ significantly outperforms previous state-of-the-art methods under two different settings by 2.1\% and 2.6\% respectively. 

\end{abstract}

\section{INTRODUCTION}

Scene graphs represent various entity nodes and relations among nodes as edges in the data format of graphs\cite{sg}. Entities and their relations constitute \texttt{<subject-predicate-object>} triplets in scene graphs. Scene graphs could help perform tasks related to visual understanding\cite{image_caption1,image_caption2,grounding1,vqa1,vqa2}. 
Furthermore, dynamic scene graph generation delivers frame-level scene graphs for video clips. The temporal information in dynamic scene graphs could be used for dynamic visual understanding\cite{vsg1,vsg2} and particularly conducts the decision-making process and task planning of robots particularly\cite{robot-robot-programming,robot-hierarchical,robot-self-supervised,robot-taskograph,robot-reasoning,robot-sequential}. Compared to methods of image scene graph generation, dynamic scene graph generation methods focus on the modeling of temporal information\cite{sttran,trace,meta}.

\begin{figure}[htbp]
    \centering
\includegraphics[width=\columnwidth]{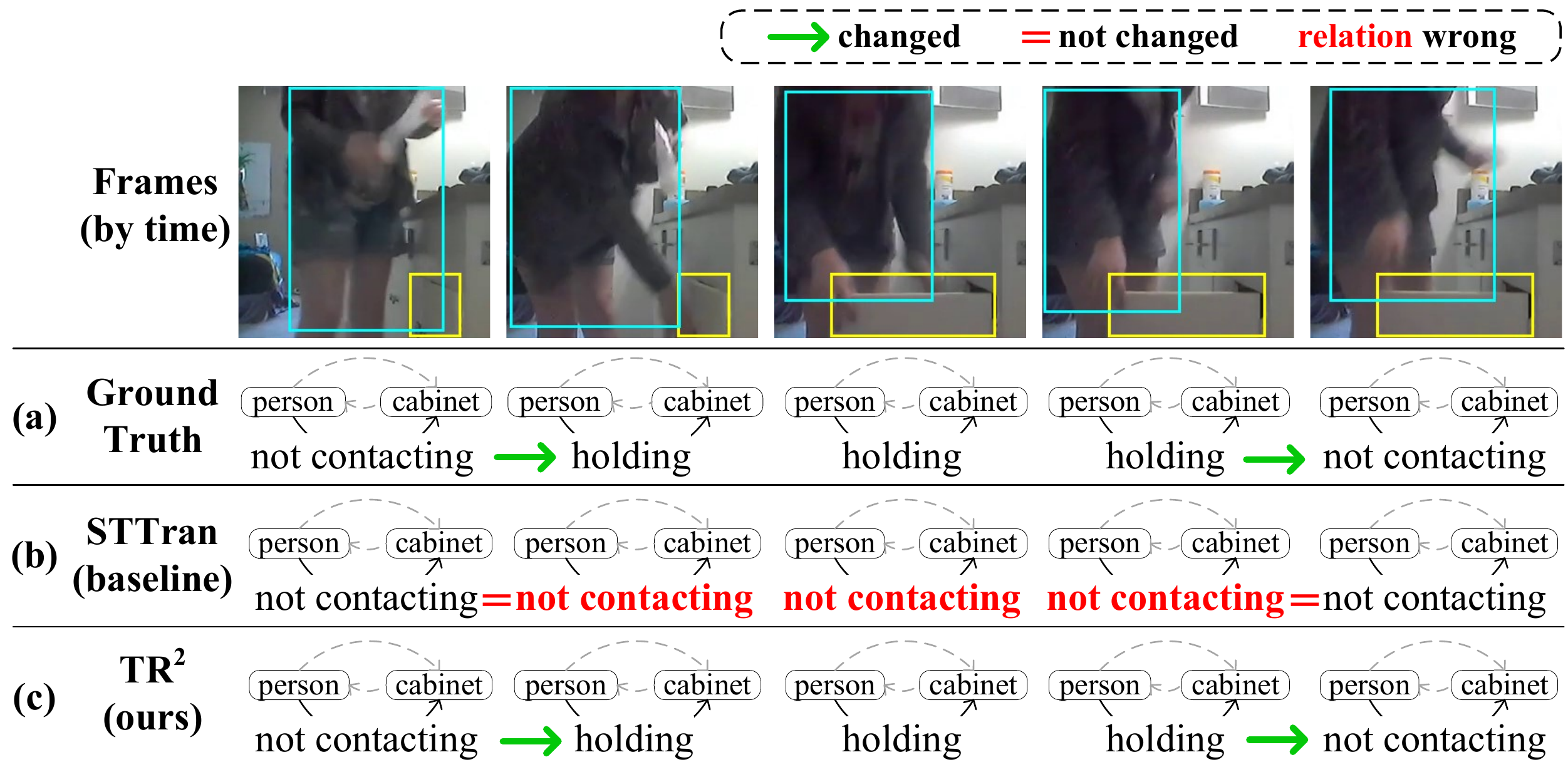}
    \caption{An example of the existing method fails to judge the change of relations in dynamic scene graphs. The person is bounded by blue boxes and the cabinet is bounded by yellow boxes. The person is opening the cabinet from the second frame to the fourth one and withdraws her hand at the fifth frame eventually. (a) The ground truth that represents the temporal change of the relation between the person and the cabinet. (b) The generation results obtained by STTran\cite{sttran}, where the same relation and scene graphs are retained by mistake. (c) Our TR$^2$ succeeds in judging the change of relations in the second frame and the fifth frame.}
    \label{fig-motivation}
\end{figure}

There are significant advances in the area of dynamic scene graph generation in recent years, where the learning of relation features is exploited to perform relation classification. However, the existing methods behave ambiguously when they should judge if the relation between a \texttt{subject-object} pair differs from that in the last frame. 
On the one hand, the subtle movement of entities may imply the change of relation between frames. However, the subtle change would be hard for the visual backbone to recognize. Therefore, the existing methods prefer maintaining the same relations with the former frames. For example, in Fig. \ref{fig-motivation} (b), the existing method STTran\cite{sttran} fails to capture the change of contacting relations between the person and the cabinet and insists that the person is not contacting the cabinet. 
On the other hand, sometimes the form or the position of entities changes a lot compared with the last frame but the corresponding relations remain the same in fact. Encountering such a situation, the existing methods may be confused by the obvious movement of entities. After that, the existing methods predict wrong relations with fake changes.

Existing methods have poor performance on the judgment of the change of relations because of their negligence in the modeling of the difference of relation features in adjacent frames. Without full use of the labels and information about relation changes that are implied in the dataset, existing methods are judging relations of each frame independently despite the temporal information fusion\cite{sttran,trace}.

To address this problem about the ambiguity for time-variant relations, the improvement should focus on the modeling and guidance of the temporal difference of relation features, which corresponds to the difference of relation labels. Accordingly, we propose a Time-variant Relation-aware TRansformer (TR$^2$) for dynamic scene graph generation. 
Explicitly, TR$^2$ extracts the difference of relation features in adjacent frames and constrains it with the situation of change of corresponding relations. We perform cross-modality knowledge distillation for the learning of time-variant relations. Furthermore, TR$^2$ guides the difference of relation features with text embeddings of the prompted sentences like "a photo of a \texttt{subject} \texttt{predicating} an \texttt{object}" labelled with text relation. 
Implicitly, the relation feature fusion module in TR$^2$ performs intra-frame and inter-frame information fusion with a transformer. Besides, TR$^2$ emphasizes the relation change with a message token which takes the influence degree of the last frame on the current frame into account. 
With the above explicit and implicit modeling of the time-variant relations, TR$^2$ could judge the temporal modification of scene graphs correctly.

We evaluate our TR$^2$ on the  Action Genome (AG) benchmark\cite{ag}. Extensive experiments demonstrate the effectiveness of our explicit and implicit modeling of time-variant relations. Our main contributions are summarized as follows:

\begin{enumerate}
    \item For the first time, cross-modality guidance is performed for time-variant relations in dynamic scene graph generation. We use the difference of text embeddings of prompted sentences about relation labels as the supervision signal for relations.
    \item We design a relation feature fusion module with a message token to model the relation features and their temporal differences implicitly.
    \item We propose a new framework called Time-variant Relation-aware TRansformer (TR$^2$), which achieves new state-of-the-art performances on the public AG dataset for dynamic scene graph generation.
\end{enumerate}

\section{RELATED WORK}
\subsection{Scene Graph Generation}
A scene graph expresses the entities and their relations in a single image in a graphical structure where nodes indicate the entities and edges indicate relations among the entities\cite{sg,sg7}. 
\cite{sg1} and \cite{sg2} jointly consider local visual features and introduce a box attention mechanism. The interactions between local features are used to improve scene graph generation (SGG) performance\cite{sg3,sg4}. \cite{sg5} found the strong regularization for relationship prediction provided by statistical co-occurrences of the Visual Genome dataset\cite{sg6}.

Based on image SGG, Ji et al.\cite{ag} released the dataset Action Genome which is the benchmark dataset for video SGG now. Video SGG methods provide frame-level or clip-level scene graphs\cite{feng2021exploiting}. \cite{sttran} encodes the spatial context within single frames and decodes with a temporal module. \cite{trace} proposed a hierarchical relation tree and aggregates the context information efficiently with the tree. \cite{tracklet1, tracklet2} generate clip-level scene graph generation based on tracklet computations. \cite{meta} handles the biases in Video SGG with the help of meta learning.

\subsection{Scene Graphs for Robot Planning}
Scene graph representations naturally express objects and predicates, which provides feasibility for task planning of robots\cite{robot-sequential}. Scene understanding helps various visual tasks\cite{icra1,icra2,icra3,icra4,icra5,icra6}. Therefore, robot planning with scene graphs attracts great attention in recent years\cite{robot-hierarchical,robot-taskograph,robot-self-supervised}. For example, \cite{robot-robot-programming} presents a goal-directed Programming by Demonstration system at the level of scene graphs that focuses on the poses of objects and considers the robot as an operator in the scene. 
\cite{robot-reasoning} pays attention to robot planning under partial observability and uses local scene graphs of single images to build and augment global scene graphs toward context-aware robot planning under partial observability. 
Recently, \cite{robot-sequential} demonstrates the graph-based planning scheme in complex sequential manipulation tasks by attaching extra task-dependent attributes information on nodes as attributes to constrain the possible interactions with the node.

\subsection{Knowledge Distillation}
The guidance on the relation features in our TR$^2$ model is inspired by knowledge distillation. Knowledge distillation is first proposed in \cite{kd1} which means distilling the knowledge from a larger deep neural network into a small network. 
Feature-based knowledge distillation uses the output of intermediate layers, i.e. feature maps as the knowledge to supervise the training of the student model\cite{kd-feature-1,kd-feature-2,kd-feature-3,kd-feature-4}. 
Most of the previous knowledge distillation methods work offline\cite{kd1,kd-offline-1,kd-offline-2,kd-offline-3,kd-offline-4} and so does our TR$^2$ model. In offline knowledge distillation, the teacher model is trained on a large set of training data before distillation. The intermediate features of the large teacher model are distilled into the student model to guide its training process.
Cross-modal distillation transfers knowledge between different modalities. For example, \cite{kd-offline-2} proposed the probabilistic knowledge distillation and transferred knowledge from the textual modality into the visual modality. 

\begin{figure*}[htbp]
\vspace*{5pt}
    \centering
\includegraphics[width=\textwidth]{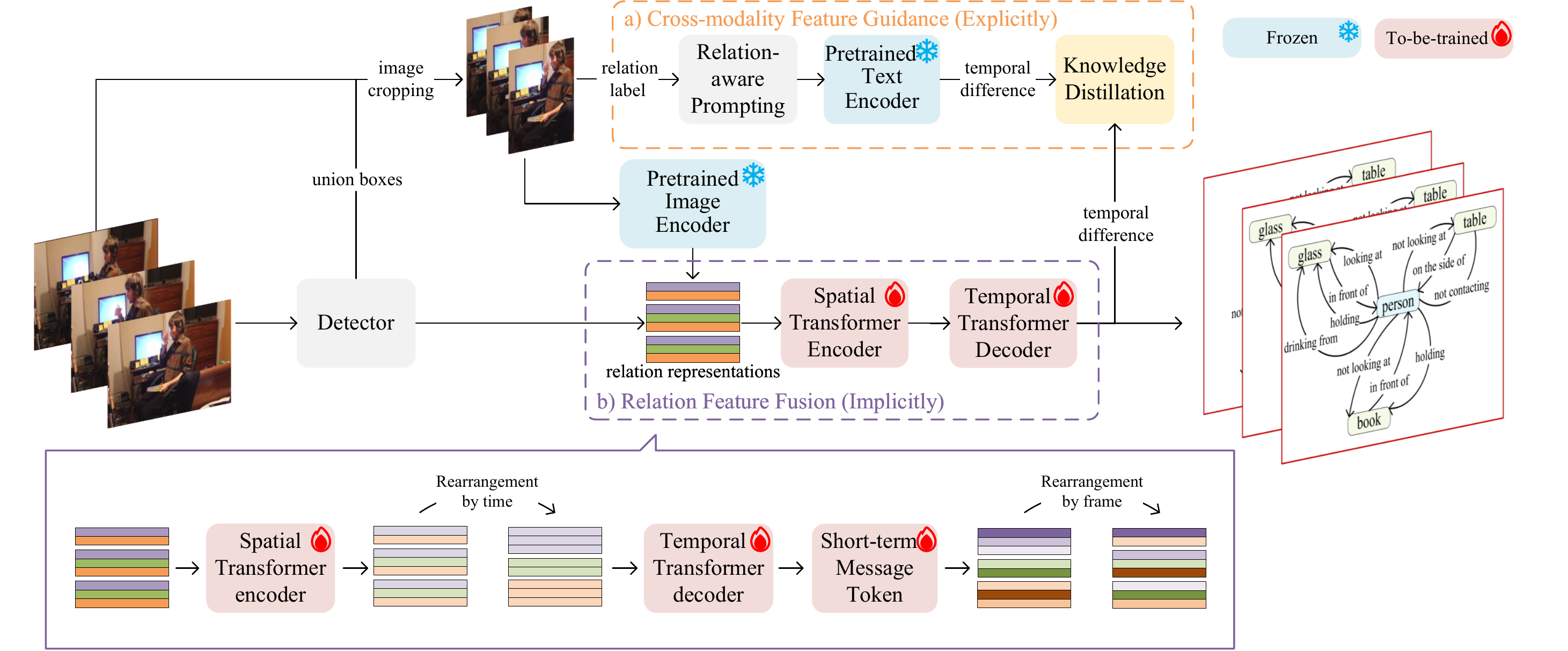}
    \caption{Our TR$^2$ framework. The detector provides the bounding box, the prediction of categories, and visual features of entities in frames, respectively. The bounding boxes of the subject and the object in a pair are combined as the union bounding box, which is used to crop the original frames to get the region of interest. The cropped images are fed into the pretrained image encoder and become relation representation components in order to adapt the cross-modality distillation process. The guidance module in the orange dotted box a) is applied in the training phase only. The spatial and temporal transformer module in purple boxes b) perform intra-frame and inter-frame information fusion on the relation representation. The short-term message token emphasizes the influence of last frames on current frames. Parameters of modules with the fire symbol are updated during training and that of modules with the snow symbol are frozen. In the relation feature fusion module, colored lines in a block indicate an input unit of the transformer module. Lines in the same or similar color mean that they are representations of the same entity in different frames. Best viewed in color.}
    \label{fig-model}
\end{figure*}

\section{Time-variant Relation-aware Transformer}
Let $G=\left\{ G_t \right\}_{t=1}^T$ denote dynamic scene graphs of a video clip, where $t$ indicates the frame index and $T$ stands for the total number of labeled frames in the video. Suppose $\left\{ i_t \right\}_{t=1}^T$ expresses the frames in the video. $G_t = \left\{ V_t, E_t \right\}$ stands for the scene graph of frame $i_t$, where $V_t$ is the set of entities as nodes and $E_t$ is the set of relations as edges among nodes in $V_t$ of frame $i_t$. Entities in $V_t$ and relationships in $E_t$ form multiple \texttt{<subject-predicate-object>} triplets. With the images of keyframes in a video as the input, our TR$^2$ model produces the corresponding dynamic scene graphs as the output. In the following, we introduce the overall model framework of TR$^2$ and then clarify its key modules.

\subsection{The Overall Framework}
In this paper, we propose 
TR$^2$ for dynamic scene graph generation. Fig. \ref{fig-model} shows the overall framework of our TR$^2$ model.

First, the frames are fed into an object detector. The detector detects entities in frames and provides the bounding boxes, categories, and visual features of the detected entities. The visual features from the detector serve as a component of the representations of relations in frames, i.e., the representations of \texttt{predicate} in \texttt{<subject-predicate-object>} triplets. We combine the detected bounding boxes of the \texttt{subject} and the \texttt{object} in \texttt{<subject-predicate-object>} triplets as union boxes.

Second, we crop the input images with the union boxes to keep the regions of interest that are corresponding to the triplets in the union boxes. We use the image encoder of a large-scale vision-and-language model to encode the cropped images to adapt properly to the guidance given in \ref{kd}. The prompted relation labels of the cropped image would participate in the cross-modality feature guidance.

Third, the relation representations obtained in the last two steps would be fed into the relation feature fusion module. The fusion module detailed in \ref{transformer} carries out intra-frame and inter-frame feature fusion. The relation features after information fusion are classified and form scene graphs with the entity nodes detected by the object detector. In the training phase, the output features of the relation feature fusion module are guided in the cross-modality feature guidance module that would be in details in \ref{kd}.

\subsection{Relation Feature Fusion}
\label{transformer}
The purple box b) in Fig. \ref{fig-model} shows the relation feature fusion module based on a spatial-temporal transformer. 
First, TR$^2$ performs intra-frame spatial feature fusion on the relation representations. The spatial encoder operates among relations in a frame without position embedding. 
Second, we rearrange the relation representations by entity and time before the temporal module. In this step, the relation features of a pair of entities in all frames that the pair appears are gathered. 
Then, the temporal decoder performs long-term inter-frame fusion on the temporal sequence of relation features of each entity pair with temporal position embedding. 
After that, we model the time-variant relations implicitly with a short-term message token. Specifically, features are calculated with the message token like
\begin{equation}\label{equ-token}
    e_{r_t} =  {\rm Concat} \left ( e_{f_t} , e_{f_{t-1}} \cdot m_{t-1}  \right )
\end{equation}
where $e_f$ stands for the output of the temporal decoder. $t$ and $t-1$ in the subscripts indicate the current frame and the last frame. $m_{t-1}$ is our short-term message token that evaluates how much the last frame affects the current frame. The message token is calculated with $m_{t-1}=g( {\rm Concat}( e_{f_t}, e_{f_{t-1}} ) )$ where $g$ is a feed-forward network. Then we concatenate the product of the message token $m_{t-1}$ and the features of the last frames to the features of the current frames. $e_r$ denotes the concatenate result. 
At last, the relation representations are rearranged by frame again to facilitate the scene graph output for each frame.

\subsection{Cross-Modality Feature Guidance}
\label{kd}


As we mentioned before, we perform cross-modality feature guidance on relation features in adjacent frames explicitly. In this way, we alleviate the problem that existing methods behave ambiguously when dealing with time-variant relations. 
To be specific, we use the knowledge of a pretrained large-scale vision-and-language model as the supervision signal to guide TR$^2$ besides the original target of scene graph labels. Different from the original feature-based knowledge distillation, we perform temporal difference on the guiding and guided features before distillation. We clarify the specific guidance process as follows.

\begin{figure}[htbp]
\vspace{5pt}
    \centering
\includegraphics[width=\columnwidth]{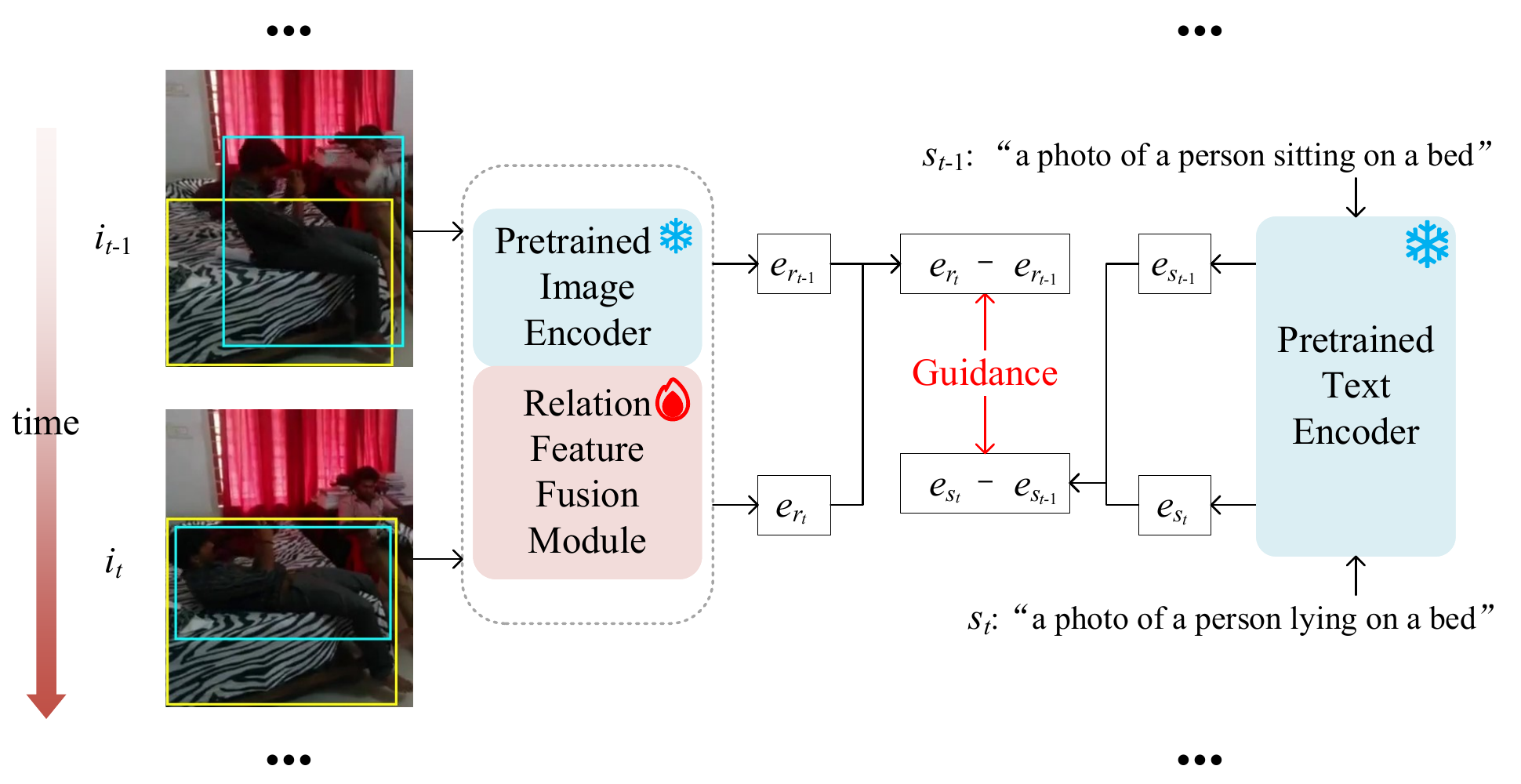}
    \caption{The cross-modality feature guidance module. We use a pretrained text encoder to extract the text embeddings of prompted sentences that describe the relations in scene graphs. For every two adjacent labeled frames, we perform knowledge distillation from the temporal difference of text embeddings to the temporal difference of relation features. Parameters of the pretrained image and text encoders are frozen.}
    \label{fig-kd-mini}
\end{figure}

The orange box a) in Fig. \ref{fig-model} shows the input of the cross-modality guidance module and the specific process is shown in Fig. \ref{fig-kd-mini}. $i_{t-1}$ and $i_t$ in Fig. \ref{fig-kd-mini} are two adjacent labeled frames. 
After the relation feature fusion module, we get the representations of the relation between the person and the bed, denoted by $e_{r_{t-1}}$ and $e_{r_t}$ for $i_{t-1}$ and $i_t$, respectively. At the same time, we mine the information in the text embeddings of prompted text labels of time-variant relations. To be specific, we extract the words of the subjects, objects, and their relations in scene graphs. Using these words, we construct the sentence of "a photo of a \texttt{subject} \texttt{predicating} an \texttt{object}" with "a photo of" as the prompt. For example, the prompted sentence for $i_{t-1}$ is "a photo of a person sitting on a bed" and the prompted sentence for $i_t$ is "a photo of a person lying on a bed". Feeding the prompted sentences into the text encoder of the pretrained large-scale vision-and-language model, we get the text features of the sentences denoted by $e_{s_{t-1}}$ and $e_{s_t}$. 
With relation features after the relation feature fusion module and text embeddings of prompted sentences, we perform temporal difference and obtain $e_{r_t}-e_{r_{t-1}}$ and $e_{s_t}-e_{s_{t-1}}$. TR$^2$ employs $e_{s_t}-e_{s_{t-1}}$ as the supervision signal to guide $e_{r_t}-e_{r_{t-1}}$. Similar operations are performed on every two adjacent frames. In this way, the teacher model helps TR$^2$ to be sensitive to time-variant relations. As a result, TR$^2$ alleviates the ambiguity towards the temporal change of relations.

\subsection{Training}
The training objective function consists of three components, i.e., the entity classification loss, the relation classification loss, and the knowledge distillation loss.

First of all, after the pretrained object detector, TR$^2$ utilizes cross-entropy loss to measure the prediction of entity classification. We utilize $L_{obj}$ to denote this entity loss term. 
Second, as for the classification of relations among entities, we use binary cross-entropy loss to deal with the multi-label predicates through treating the judgment of each label as binary classification. After that, we gather them in a focal loss form. We adopt $L_{rel}$ to express this relation loss term. 
Third, in the training phase, cross-modality feature guidance is done. We exploit mean-squared loss for the distillation here, i.e.,
\begin{equation}
    L_{guidance}=\frac{1}{T-1} \sum_{i=2}^{T} \left [ \left ( e_{r_t}-e_{r_{t-1}} \right ) - \left ( e_{s_t}-e_{s_{t-1}} \right )   \right ]  ^ { 2 }.
\end{equation}
Then the total loss can be formulated as
\begin{equation}
    L= \lambda  L_{obj} + L_{rel}+L_{guidance}
\end{equation}
where $\lambda$ is the weight of the entity detection loss that distinguishes the detection stage from the relation learning stage.

\section{EXPERIMENTS}
We present the experimental results of TR$^2$ on dynamic scene graph generation. First, we introduce the dataset preparation and experiment settings. Second, the implementation details of TR$^2$ in experiments are provided. Finally, we show and analyze the experimental results that include but are not limited to overall assessments and ablation studies. We present some visualization cases in the accompanying video.

\subsection{Experimental Setup}

\subsubsection{Dataset}
Our experiments are conducted on the AG dataset\cite{ag}, which is the benchmark dataset of dynamic scene graph generation. AG is built upon the Charades dataset\cite{charades} and provides frame-level scene graph labels with a total of 234,253 frames in 9,848 video clips. In AG, there contain 36 types of entities and 26 types of relations in the label annotations. Such 26 types of relations are divided into three classes, i.e., attention, spatial, and contacting relations. The attention relations are used to describe if a person is looking at an object or not. The spatial relations specify the relative position. The contacting relations represent different ways of contacting in particular.

\begin{table*}[t]
\vspace*{5pt}
\centering
\caption{Recall (\%) comparison results of our TR$^2$ and baselines in the With Constraints and the No Constraints settings}
\label{table-constraint}
\tabcolsep=0.06cm
\begin{tabular}{ccccccccccccccccccc}
\toprule
\multirow{3}{*}{Method} & \multicolumn{9}{c}{With Constraints}  & \multicolumn{9}{c}{No Constraints}    \\ 
\cmidrule(r){2-10} \cmidrule(r){11-19}
& \multicolumn{3}{c}{PredCls} & \multicolumn{3}{c}{SgCls} & \multicolumn{3}{c}{SgDet} & \multicolumn{3}{c}{PredCls} & \multicolumn{3}{c}{SgCls} & \multicolumn{3}{c}{SgDet} \\ 
\cmidrule(r){2-4} \cmidrule(r){5-7} \cmidrule(r){8-10} \cmidrule(r){11-13} \cmidrule(r){14-16} \cmidrule(r){17-19}
& R@10    & R@20    & R@50    & R@10    & R@20   & R@50   & R@10    & R@20   & R@50   & R@10    & R@20    & R@50    & R@10    & R@20   & R@50   & R@10    & R@20   & R@50   \\ 
\midrule
M-FREQ\cite{motif}\tiny{\emph{CVPR'18}} & 62.4 & 65.1 & 65.1 & 40.8 & 41.9 & 41.9 & 23.7 & 31.4 & 33.3 & 73.4 & 92.4 & 99.6 & 50.4 & 60.6 & 64.2 & 22.8 & 34.3 & 46.4 \\
RelDN\cite{reldn}\tiny{\emph{CVPR'19}} & 66.3 & 69.5 & 69.5 & 44.3 & 45.4 & 45.4 & 24.5 & 32.8 & 34.9 & 75.7 & 93.0 & 99.0 & 52.9 & 62.4 & 65.1 & 24.1 & 35.4 & 46.8  \\
TRACE\cite{trace}\tiny{\emph{ICCV'21}}   & 64.4 & 70.5 & 70.5 &  36.2  & 37.4 & 37.4 & 19.4  & 30.5 & 34.1  & 73.3  & 93.0 & 99.5  & 36.3 & 45.5  & 51.8 & 27.5  & 36.7  & 47.5 \\
STTran\cite{sttran}\tiny{\emph{ICCV'21}} & 68.6    & 71.8    & 71.8    & 46.4    & 47.5   & 47.5   & 25.3    & 34.1   & 37.0   & 77.9    & 94.2    & 99.1    & 54.0    & 63.7   & \textbf{66.4}   & 24.6    & 36.2   & 48.8\\
Ours& \textbf{70.9} & \textbf{73.8} & \textbf{73.8} & \textbf{47.7} & \textbf{48.7} & \textbf{48.7} & \textbf{26.8} & \textbf{35.5} & \textbf{38.3}& \textbf{83.1} & \textbf{96.6} & \textbf{99.9} & \textbf{57.2} & \textbf{64.4} & 66.2  & \textbf{27.8} & \textbf{39.2} & \textbf{50.0}\\
\bottomrule
\end{tabular}
\end{table*}

\begin{table}[t]
\centering
\caption{Recall (\%) comparison results with top 6 predictions for each pair}
\label{table-top6}
\tabcolsep=0.12cm
\begin{tabular}{ccccccc}
\toprule
\multirow{3}{*}{Method} & \multicolumn{2}{c}{PredCls} & \multicolumn{2}{c}{SgCls} & \multicolumn{2}{c}{SgDet}  \\ \cmidrule(r){2-3} \cmidrule(r){4-5} \cmidrule(r){6-7} 
 & R@20    & R@50   & R@20   & R@50   & R@20   & R@50  \\ \midrule
M-FREQ\cite{motif}\tiny{\emph{CVPR'18}} & 85.9 & 89.4  & 44.9  & 47.2 & 34.5 & 43.7 \\
RelDN\cite{reldn}\tiny{\emph{CVPR'19}} &89.6  & 93.6 & 46.8 & 49.1   & 35.2  & 44.9 \\
TRACE\cite{trace}\tiny{\emph{ICCV'21}} &  90.8 & 94.0 & 48.1 & 50.3  & 37.3 & 47.4  \\
STTran\cite{sttran}\tiny{\emph{ICCV'21}} &90.2  &92.1 &60.6  &61.4  &36.0   &47.2 \\
MVSGG\cite{meta}\tiny{\emph{ECCV'22}} & 90.5 &94.1  &47.7 & 50.0 & 36.8  & 46.7  \\
Ours & \textbf{93.5} & \textbf{95.0} & \textbf{62.4} & \textbf{63.1}  & \textbf{39.1} & \textbf{48.7} \\
\bottomrule
\end{tabular}
\vspace{-0.3cm}
\end{table}

\subsubsection{Evaluation Tasks and Metrics} 
In the same way as \cite{ag,trace}, we make the evaluation of TR$^2$ on the AG dataset under three tasks below: predicate classification (PredCls), scene graph classification (SgCls), and scene graph detection (SgDet). In PredCls, the bounding boxes and object categories are provided and the model needs to predict predicate categories. In SgCls, only the bounding boxes of entities are given. In SgDet, the model needs to detect entities and predict the predicates. In line with \cite{sttran,trace}, $Recall@K$ (i.e. $R@K$, $K=[10,20,50]$) is adopted as the evaluation metric. As for the predicate choice of predictions for each pair, we borrow the settings from both the \textbf{With Constraints} and the \textbf{No Constraints} strategies from \cite{sttran} and \textbf{top $k$ predictions} ($k=6$) from \cite{trace} to make fair and sufficient comparison with baselines. In these predicate choice settings, \textbf{With Constraints} is the most stringent since it only chooses one predicate for each entity pair. \textbf{No Constraints} allows multiple predictions of relations for each entity pair taking top $100$ predicates for all pairs in a single frame. Eclectically, \textbf{top $k$ predictions} picks the first $k$ predictions sorted by scores for each pair.

\subsection{Implementation Details}
The object detector is implemented with a Faster-RCNN\cite{fasterrcnn} network based on ResNet-101\cite{resnet101}. 
As for the relation feature fusion module, the spatial module is a $1$-layer encoder and the temporal module is a $3$-layer decoder with 8 heads. The feed-forward dimension is $2048$ and the dropout is $0.1$ in this transformer. 
In the cross-modality guidance module, we use CLIP which is a large-scale pretrained vision-and-language model to obtain the text embeddings of prompted sentences. Specifically, we use the VIT-B-32 model provided by CLIP\footnote{https://github.com/openai/CLIP}.
We adopt the AdamW optimizer\cite{adamw} to train our TR$^2$ with the initial learning rate of $1e^{-5}$.


\subsection{Overall Performance Assessments}

The comparison results of TR$^2$ and baselines are summarized in Table \ref{table-constraint} and Table \ref{table-top6}, respectively. In Table \ref{table-constraint}, we report the performance of TR$^2$ in the settings of \textbf{With Constraints} and \textbf{No Constraints}. The values of $R@20$ and $R@50$ of PredCls task in \textbf{With Constraints} are almost the same because this setting only chooses one predicate for each pair so there are a few cases that could reach 50 predictions in total.
In Table \ref{table-top6}, we report the results under the setting of \textbf{top $k$ predictions} with $k=6$. 

Analyzing the experimental results in Table \ref{table-constraint} and Table \ref{table-top6}, it is obvious that TR$^2$ achieves the new state-of-the-art results on almost all the metrics and settings. Compared to M-FREQ\cite{motif} and RelDN\cite{reldn}, the effect of the basic temporal modeling in TR$^2$ is proved, which is also adopted in STTran\cite{sttran} and TRACE\cite{trace}. Furthermore, with the guidance on the temporal difference between adjacent frames, TR$^2$ performs better than STTran\cite{sttran} and TRACE\cite{trace}. 
In the SgCls task, the $R@50$ value of STTran looks slightly higher than that of TR$^2$ because of the inherent gap inside the detector.
The PredCls task attaches great importance to the relation classification and the temporal difference of dynamic scene graphs. In the PredCls task, TR$^2$ yields $2.1\%$ improvement over STTran\cite{sttran}, which is our main baseline, and $4.5\%$ improvement over TRACE\cite{trace} in the \textbf{With Constraints} setting. TR$^2$ outperforms previous state-of-the-art methods by $2.6\%$ in the PredCls task under the \textbf{No Constraints} setting.

\subsection{Ablation Study}
In order to confirm the effect of each module in our TR$^2$, we perform ablation experiments on different guidance settings and the relation feature fusion module. Experiments of our ablation studies are all done on the PredCls task for the AG dataset in the \textbf{With Constraints} setting.

\textbf{Guidance Settings.}
As for the guidance module, which corresponds to our explicit modeling of the time-variant relations in dynamic scene graphs, we conduct the ablation study in the following two aspects:

\begin{figure}[t]
    \centering
\includegraphics[width=\columnwidth]{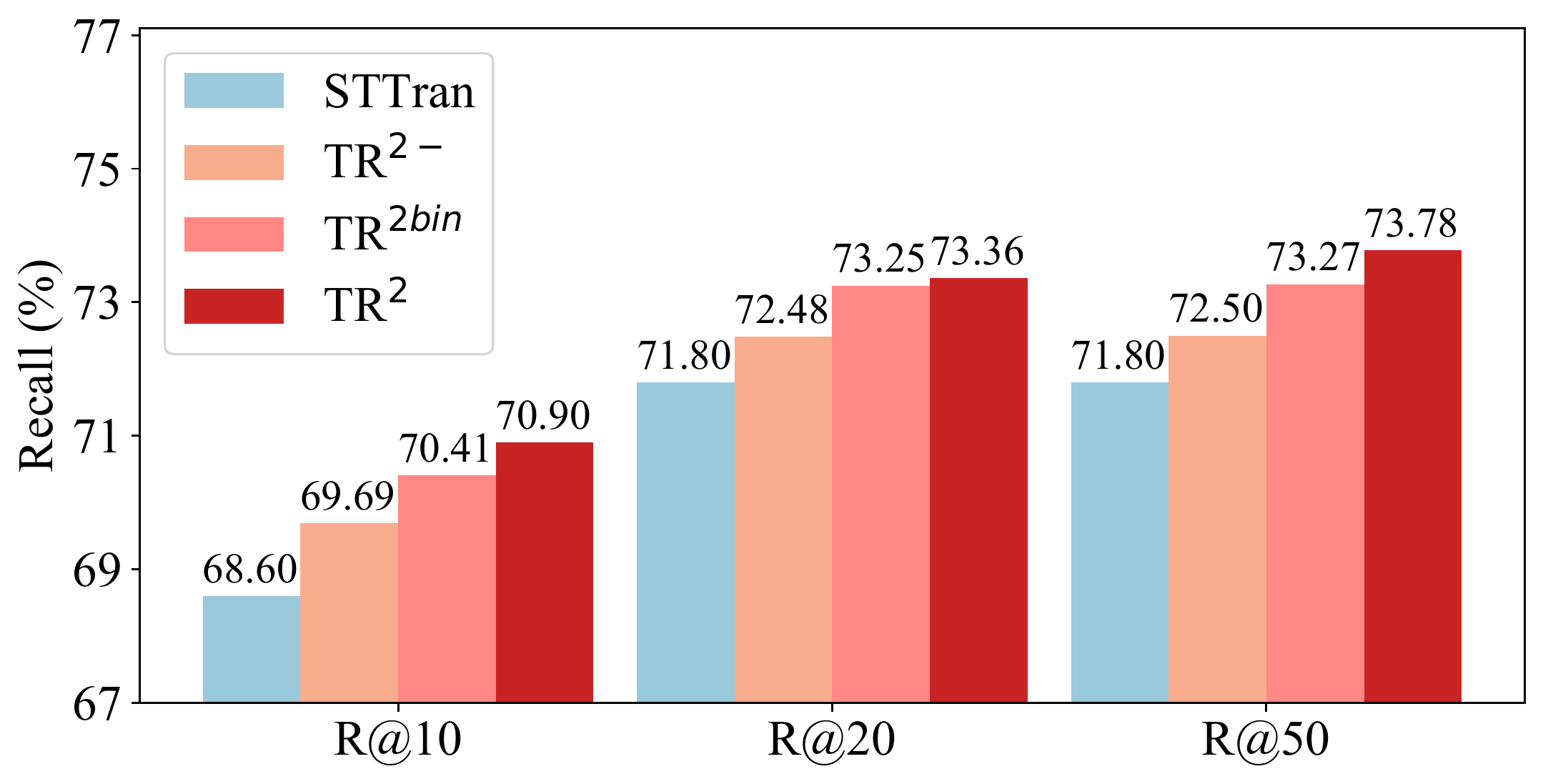}
    \caption{The ablation study of the cross-modality feature guidance module. TR$^{2-}$ means TR$^2$ without the guidance module. TR$^{2bin}$ considers the guidance as a binary classification task. TR$^2$ denotes the vanilla model and STTran\cite{sttran} is also illustrated for comparison.}
    
    \label{fig-exp-softlabel}
    \vspace{-0.3cm}
\end{figure}

\textbf{a) The effect of the guidance module.} We perform ablation studies on TR$^2$ without guidance and with simpler guidance. The results are presented in Fig. \ref{fig-exp-softlabel}. TR$^{2-}$ in Fig. \ref{fig-exp-softlabel} removes the guidance module and is supervised by the classification of relations alone. TR$^{2bin}$ considers the guidance of temporal change as a binary classification task, that is, we use the temporal difference of relation features to classify if the relation change or not. 
As shown in Fig. \ref{fig-exp-softlabel}, TR$^{2-}$ performs better than STTran\cite{sttran} owing to the temporal decoder that passes long-term information and the message token that emphasizes short-term influence. With the simple coarse-grained binary guidance, TR$^{2bin}$ outperforms TR$^{2-}$ and could model the change of relations already. Furthermore, with fine-grained prompted text embeddings that represent the relations, TR$^2$ yields better results than TR$^{2bin}$.

In addition, we tried several different kinds of prompts and the aggregated one of multiple prompts to find that different prompts make little difference. This result demonstrates that it is our leverage of the label text that improves the generation performance rather than the prompting words.

\begin{table}[t]
\vspace*{5pt}
\centering
\caption{Ablation study of temporal differences preceding guidance module}
\label{table-ablation-guidance-difference}
\begin{tabular}{cccc}
\toprule
Temporal Difference & R@10                 & R@20 & R@50 \\
\midrule
 -  & 70.1 & 73.0 & 73.0 \\
\checkmark & 70.9{$_{\textcolor{red}{(+0.8)}}$}   & 73.8{$_{\textcolor{red}{(+0.8)}}$}  & 73.8{$_{\textcolor{red}{(+0.8)}}$}  \\
\bottomrule
\end{tabular}
\end{table}

\textbf{b) The temporal difference before guidance.} The knowledge distillation module in TR$^2$ focus on the change of relations in dynamic scene graphs. We also had a trial of guidance without the temporal difference, that is, distillation on temporal items directly and separately. In this experiment, the $L_{guidance}$ is updated as $L^{'}_{guidance}$ with
\begin{equation}
    L^{'}_{guidance}=\frac{1}{T-1} \sum_{i=2}^{T} \left ( e_{r_t}-e_{s_t} \right )   ^ { 2 }.
\end{equation}
The comparison results in Table \ref{table-ablation-guidance-difference} show the importance of temporal difference in the guidance module. Guidance on the temporal difference of features is more beneficial than guiding separate temporal items directly.


\begin{table}[t]
\centering
\caption{Ablation study on the relation feature fusion module of TR$^2$. Token means the short-term message token.}
\label{table-ablation-st}
\tabcolsep=0.12cm
\begin{tabular}{ccccc}
\toprule
\multicolumn{1}{c}{Spatial} & Temporal & R@10                 & R@20 & R@50 \\
\midrule
-  & -  & 69.5 & 72.3 & 72.3 \\
- & \checkmark & 70.2{$_{\textcolor{red}{(+0.7)}}$} & 73.1{$_{\textcolor{red}{(+0.8)}}$} & 73.1{$_{\textcolor{red}{(+0.8)}}$} \\
\checkmark & -  & 69.7{$_{\textcolor{red}{(+0.2)}}$} & 72.5{$_{\textcolor{red}{(+0.2)}}$} & 72.5{$_{\textcolor{red}{(+0.2)}}$} \\
\checkmark & Decoder   & 70.5{$_{\textcolor{red}{(+1.0)}}$} & 73.3{$_{\textcolor{red}{(+1.0)}}$} & 73.4{$_{\textcolor{red}{(+1.1)}}$} \\
\checkmark & Token & 70.0{$_{\textcolor{red}{(+0.5)}}$} & 72.8{$_{\textcolor{red}{(+0.5)}}$} & 72.9{$_{\textcolor{red}{(+0.6)}}$} \\
\checkmark & Decoder+Token & 70.9{$_{\textcolor{red}{(+1.4)}}$}  & 73.8{$_{\textcolor{red}{(+1.5)}}$} & 73.8{$_{\textcolor{red}{(+1.5)}}$} \\
\bottomrule
\end{tabular}
\vspace{-0.4cm}
\end{table}

\textbf{Relation Feature Fusion Module.}
In order to illustrate the effect of each component of the relation feature fusion in TR$^2$, we conduct experiments with/without spatial or temporal modules. The results are shown in Table \ref{table-ablation-st}. According to Table \ref{table-ablation-st}, we come to the conclusion that every module related to the relation feature fusion of TR$^2$ is effective. Each module is necessary for the final best result with the $R@10$ of $70.9$.
As for the temporal modules, we ablate the decoder and the use of the message token separately. The temporal decoder could catch long-term information, while the message token emphasizes short-term information. The third row to the fifth one in Table \ref{table-ablation-st} confirm the benefit of taking long-term and short-term temporal information separately. Equipping with the long-term decoder and the short-term message token at the same time, the performance would be better as shown in the last row of Table \ref{table-ablation-st}. 

\subsection{Performance on different ratings of relation change data}
\begin{figure}[t]
\vspace*{5pt}
    \centering
\includegraphics[width=7cm]{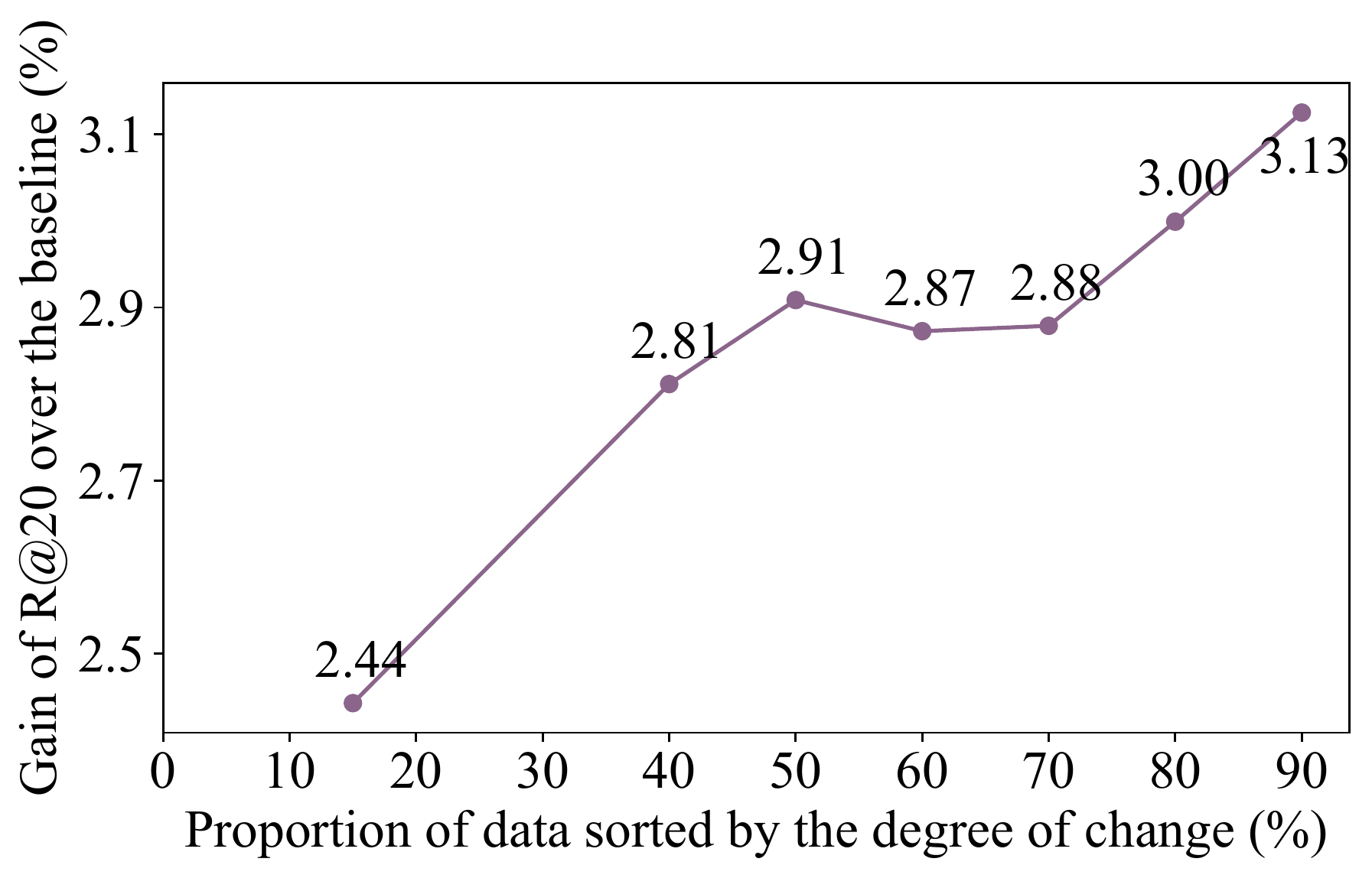}
    \caption{The more relation changes there exist in data, the more gain our TR$^2$ model outperforms the baseline (STTran).}
    \label{fig-exp-change}
\vspace{-0.4cm}
\end{figure}
As discussed above, our TR$^2$ model focuses on the change of relations between adjacent frames in dynamic scene graphs. Correspondingly, we evaluate the performance of TR$^2$ in terms of data with changes of different ratings. In particular, we sorted the data in AG\cite{ag} by the degree of change, i.e., the proportion of frames where the relations changed. The experimental results are given in Fig. \ref{fig-exp-change}, where the horizontal axis means the proportion of data sorted by change, while the vertical one indicates the gain value of $R@20$ that TR$^2$ outperforms STTran\cite{sttran}. When using data of a higher proportion, there exist more changes among frames, leading to harder difficulty for models to be tackled. The gain values of the top 30 percent of the data are combined because they keep the same relationships all the time with the degree of change as 0. Obviously, the harder the data are, the higher gain that TR$^2$ exceeds STTran\cite{sttran}. This experiment proves the superiority of TR$^2$ for modeling time-variant relations in dynamic scene graphs.

\section{CONCLUSION}
In this paper, we propose a new cross-modality time-variant relation learning method for generating dynamic scene graphs. We perform cross-modality feature guidance on the time-variant relations explicitly. We use a relation feature fusion module with a message token to model the relation features implicitly. The experimental results show that the proposed method has the best performance on the AG dataset, which outperforms existing SOTA models by 2.1\% and 2.6\% under two different settings, respectively. We perform experiments to illustrate the superiority of TR$^2$ for modeling time-variant relations in dynamic scene graphs. The interpretability of temporal modeling in dynamic scene graphs would be an interesting topic worthy of further study.

\section*{ACKNOWLEDGEMENTS}
This work was supported in part by the National Science Foundation of China (NSFC) under Grant No. 62176134, by a grant from the Institute Guo Qiang (2019GQG0002), Tsinghua University, and by research and application on AI technologies for smart mobility funded by SAIC Motor.

\clearpage

\bibliography{root}
\bibliographystyle{IEEEtran}


\end{document}